\documentclass[sigconf, screen, nonacm]{acmart}
\AtBeginDocument{%
  }

\setcopyright{acmlicensed}
\copyrightyear{2018}
\acmYear{2018}
\acmDOI{XXXXXXX.XXXXXXX}
\acmConference[Conference acronym 'XX]{Make sure to enter the correct
  conference title from your rights confirmation email}{June 03--05,
  2018}{Woodstock, NY}
\acmISBN{978-1-4503-XXXX-X/2018/06}

\acmSubmissionID{2651}

\usepackage{multirow}
\usepackage{bbding}

\begin{document}

\title{Progressive Language-guided Visual Learning for \\ Multi-Task Visual Grounding}

\author{Jingchao Wang†}
\affiliation{%
  \institution{East China Normal University}
  \city{Shanghai}
  \country{China}
}
\email{jcwang@stu.encu.edu.cn}

\author{Hong Wang†}
\affiliation{%
  \institution{Xi'an Jiaotong University}
  \city{Xi'an, Shaanxi}
  \country{China}}
\email{hongwang01@xjtu.edu.cn}

\author{Wenlong Zhang}
\affiliation{
  \institution{Shanghai AI Laboratory}
  \city{Shanghai}
  \country{China}
}
\email{zhangwenlong@pjlab.org.cn}

\author{Kunhua Ji}
\affiliation{ 
  \institution{East China Normal University}
  \city{Shanghai}
  \country{China}
}
\email{72285900034@stu.ecnu.edu.cn}

\author{Dingjiang Huang*}
\affiliation{%
  \institution{East China Normal University}
  \city{Shanghai}
  \country{China}}
\email{djhuang@dase.ecnu.edu.cn}

\author{Yefeng Zheng*}
\affiliation{%
  \institution{Westlake University}
  \city{Hangzhou, Zhejiang}
  \country{China}}
\email{zhengyefeng@westlake.edu.cn}

\thanks{† Contributed Equally.}
\thanks{* Corresponding Author.}

\renewcommand\footnotetextcopyrightpermission[1]{}
\settopmatter{printacmref=false} 
\begin{abstract}
Multi-task visual grounding (MTVG) includes two sub-tasks, \emph{i.e.}, Referring Expression Comprehension (REC) and Referring Expression Segmentation (RES). The existing representative approaches generally follow the research pipeline which mainly consists of three core procedures, including independent feature extraction for visual and linguistic modalities, respectively, cross-modal interaction module, and independent prediction heads for different sub-tasks.
Albeit achieving remarkable performance, this research line has two limitations: 1) The linguistic content has not been fully injected into the entire visual backbone for boosting more effective visual feature extraction and it needs an extra cross-modal interaction module; 2) The relationship between REC and RES tasks is not effectively exploited to help the collaborative prediction for more accurate output. 
To deal with these problems, in this paper, we propose a Progressive Language-guided Visual Learning framework for multi-task visual grounding, called PLVL, which not only finely mine the inherent feature expression of the visual modality itself but also progressively inject the language information to help learn linguistic-related visual features. In this manner, our PLVL does not need additional cross-modal fusion module while fully introducing the language guidance. Furthermore, we analyze that the localization center for REC would help identify the to-be-segmented object region for RES to some extent. Inspired by this investigation, we design a multi-task head to 
accomplish collaborative predictions for these two sub-tasks.
Extensive experiments conducted on several benchmark datasets comprehensively substantiate that our PLVL obviously outperforms the representative methods in both REC and RES tasks.
\href{https://github.com/jcwang0602/PLVL}{https://github.com/jcwang0602/PLVL}
\end{abstract}




\settopmatter{printacmref=false} 
\renewcommand\footnotetextcopyrightpermission[1]{} 
\pagestyle{plain} 

\maketitle

\section{Introduction}
\label{sec:intro}

Visual Grounding seeks to identify a visual object in an image based on a natural language expression. According to the granularity of the prediction, it can be divided into two sub-tasks: Referring Expression Comprehension (REC) and Referring Expression Segmentation (RES).
The former focuses on regional alignment with the natural language representation to generate a bounding box, and the latter aligns with the natural language representation at the pixel level, producing a segmentation mask. Recently, some researchers~\cite{su2023vglaw,cheng2023pvd,chen2024efficient} proposed a multi-task collaborative learning framework to unify REC and RES, namely Multi-Task Visual Grounding (MTVG), which proved that multi-task collaboration has much greater potential.

\begin{figure} 
    \centering  
    \includegraphics[width=1\linewidth]{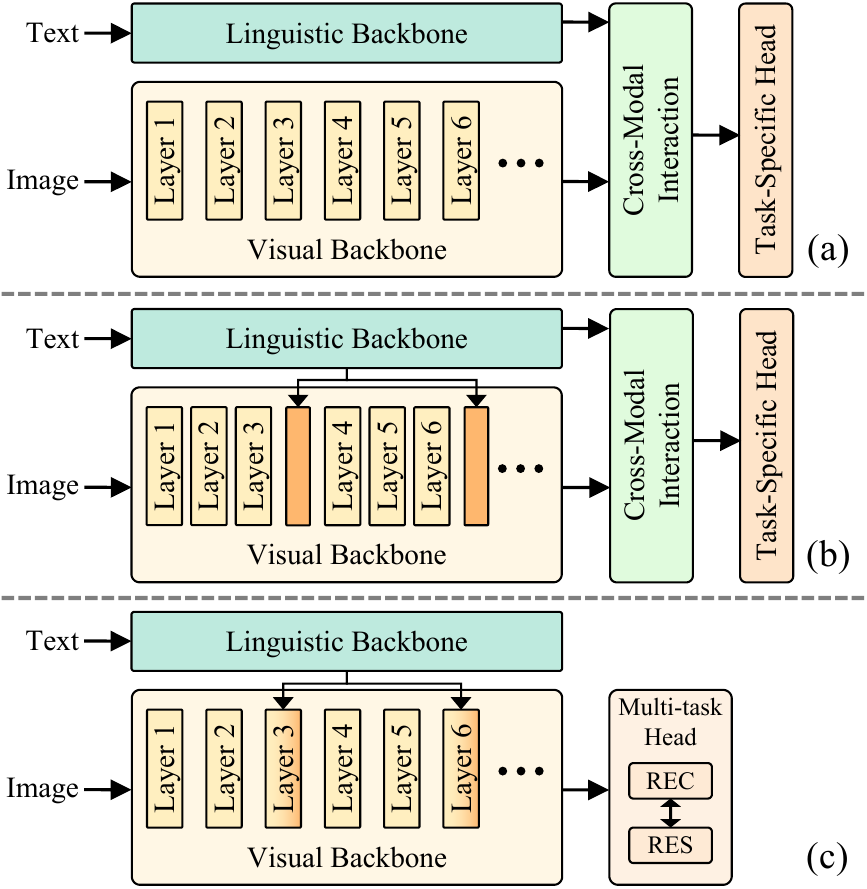}  
    \caption{Comparison of different pipelines for multi-task visual grounding: (a) Visual and language features are extracted separately and then cross-modal fusion is performed; (b) Additional modules (marked by orange) are inserted after the original network layers to inject the language features into the visual backbone; (c) Our progressive language-guided visual learning framework with a collaborative multi-task head, which directly adjusts the original network layer for progressively introducing the language guidance.}       
    \label{fig:pipeline}  
\end{figure}

For visual grounding, a prevalent research paradigm first utilizes visual backbone and linguistic backbone to extract features from image and text modalities, respectively, and then adopts a Transformer encoder/decoder for cross-modal feature fusion, as shown in Fig.~\ref{fig:pipeline}(a). While this approach has demonstrated promising performance~\cite{clipvg,chen2024efficient,segvg}, it is often hindered by a potential limitation that the linguistic-information is not injected into the visual backbone. As a result, the visual features extracted may not align well with the semantics of the natural language expression~\cite{qrnet}. Against this issue, researchers~\cite{qrnet,lgrnet} have proposed different strategies to inject the language features into the visual backbone by introducing extra network modules in order to help the extracted visual features match referring expression as much as possible, as displayed in Fig.~\ref{fig:pipeline}(b). However, an extra cross-modal interaction module still needs to be carefully designed for achieving better performance. Besides, as illustrated in Figs.~\ref{fig:pipeline}(a)(b), a common limitation of these two types of pipelines is that they use two independent heads to predict the bounding box and segmentation task, respectively, and do not take the prediction correlation between REC and RES into account, leaving room for performance improvement, especially for MTVG. To deal with these problems, very recently, a state-of-art method, called VG-LAW~\cite{su2023vglaw}, has been designed, which proposed a language adaptive weight generation strategy to help achieve the language-related visual feature extraction without introducing any extra cross-modal interaction procedure. Albeit obtaining the promising performance, it needs to generate the weights for each layer of the visual backbones, causing a certain amount of computational expense. Besides, it only utilized the class token to simultaneously guide the predictions of REC and RES tasks, but did not fully explore the cooperative relationship of these two sub-tasks from the perspective of prediction outputs.

Against these aforementioned issues, in this paper, for multi-task visual grounding (MTVG), to achieve accurate predictions based on the referring expression, it is crucial to extract effective visual features that accurately represent the objects to be identified and align with the expression cues. Based on such understandings, inspired by ViTDet~\cite{li2022vitdet}, we propose a Progressive Language-guided Visual Learning (PLVL) backbone, which mainly consists of local and global blocks. Within the local block, we utilize a self-attention mechanism to explore the relationships within the visual modality, facilitating the extraction of meaningful feature representations. Different from ViTDet, in the global block, we progressively incorporate the guidance of linguistic tokens into the visual backbone by inserting a simple cross-attention interaction operation into the original network layer (see the gradient orange module in Fig.~\ref{fig:pipeline} (c)), which aids in learning effective visual features that align with the referring expression. 
By employing this progressive language-guided visual learning approach, our framework is designed to effectively capture valuable visual features, ultimately enhancing the accuracy of predictions.
Furthermore, we recognize that the predictions of Referring Expression Comprehension (REC) and Referring Expression Segmentation (RES) have the similar central position of to-be-identified objects. Instead of adhering to the traditional method of employing two independent heads, driven by the inherent bias priors of convolutional layers,  
we introduce a pure convolution-based collaborative multi-task head that builds the bridge between REC and RES and accomplishes the joint prediction of different granularity information, thereby improving the prediction accuracy for both tasks. In summary, our contributions are mainly three-fold:

\begin{itemize}

\item  We specifically propose a progressive language-guided visual learning framework for multi-task visual grounding, which fully injects the linguistic content of referring expression into the entire visual backbone for helping more effective visual feature extraction.

\item  We deeply explore the relationship between REC and RES, and propose a novel collaborative multi-task head to carefully couple REC and RES tasks to achieve the accurate joint prediction.

\item Extensive experiments conducted on several benchmark datasets comprehensively demonstrate the effectiveness of our proposed PLVL, which always achieves superior performance on the REC task and the RES task.

\end{itemize}

\section{Related Work}
\label{sec:relatedwork}

\begin{figure*}[!t] 
    \centering  
    \includegraphics[width=0.98\linewidth]{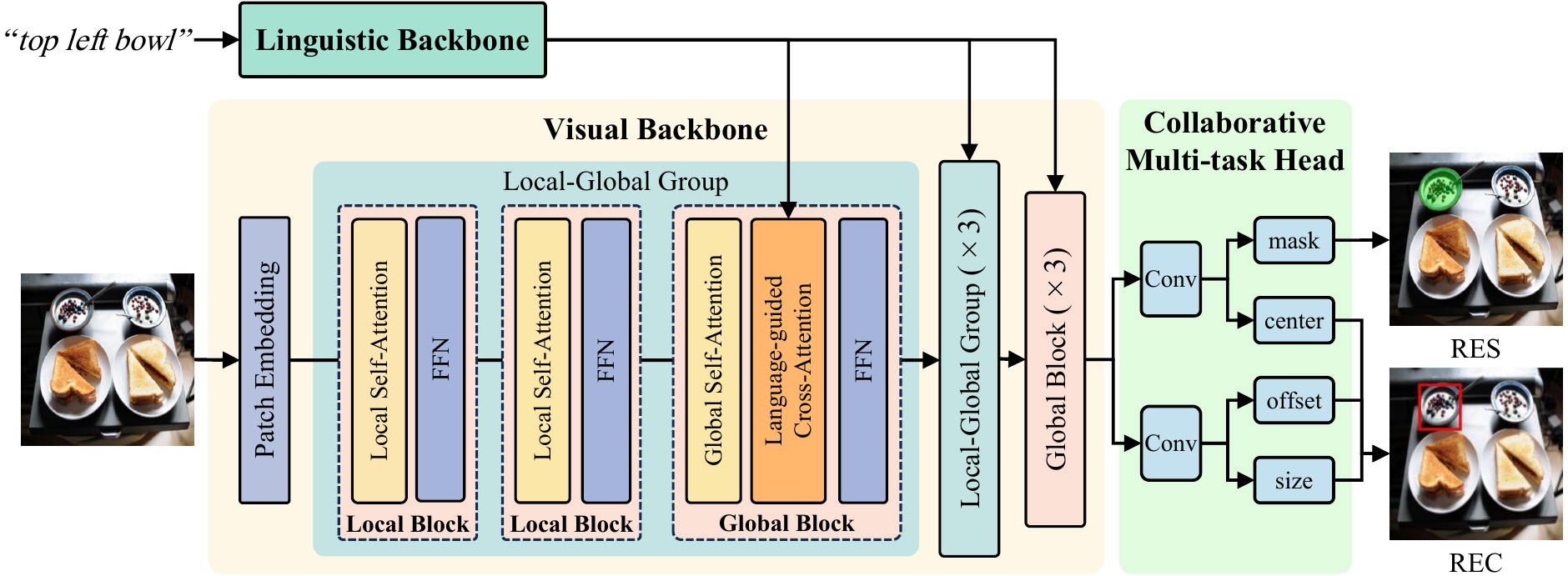}  
    \vspace{-3mm}
    \caption{Flowchart of the proposed Progressive Language-guided Visual Learning (PLVL) framework which consists of three parts, \emph{i.e.}, linguistic backbone, language-guided visual backbone, and a collaborative multi-task head. The detailed structures of local block, glocal block, and multi-task head are described in Fig.~\ref{fig:fusion}(a), Fig.~\ref{fig:fusion}(b), and Fig.~\ref{fig:head}, respectively.}       
    \label{fig:plvl}  
\end{figure*}

\subsection{Referring Expression Comprehension (REC)}

The traditional referring expression comprehension methods are generally divided into one-stage method and two-stage methods. 
The core idea of the two-stage Visual Grounding method~\cite{hu2017modeling, zhang2018grounding, hong2019learning} is to first use a pre-trained object detector to generate candidate region proposals and then select the region proposal that best matches the natural language expression as the final result.
One-stage methods\cite{yang2019fast,yang2020improving,liao2020real} typically perform cross-modal information fusion after feature extraction and then directly predict bounding boxes based on predefined anchors.

The emergence of the Transformer~\cite{vaswani2017attention,liu2021swin} has introduced a new paradigm for Visual Grounding. Recent works~\cite{deng2021transvg,chen2024efficient} utilize feature extraction networks to extract visual and language features, then employ a Transformer Encoder or Decoder to achieve cross-modal feature fusion. They introduce a REG Token to aggregate multimodal information, which is fed into a head (commonly using a Multi-Layer Perceptron) to predict the information of objects.

However, during our experiments, we found that this approach based on the REG Token has limitations that severely restrict the model's performance. Therefore, in this paper, for the REC task, we employed a classification and regression approach for prediction.

\subsection{Referring Expression Segmentation (RES)}

Unlike the REC task, which predicts a bounding box based on a referring expression, the RES task demands more refined results. 
The RES task predicts a mask, requiring the classification of each pixel in the image according to the referring expression. 
The existing work~\cite{huang2020referring,feng2021encoder,jing2021locate,yang2022lavt} has made great progress in cross-modal fusion methods and in achieving more accurate segmentation.

Different from the previous multi-modal fusion of vision and language in the network decoding stage, EFN~\cite{feng2021encoder} innovatively proposed an encoder fusion network, which realizes the gradual refinement of multimodal features, and promotes the consistent representation of cross-modal information in semantic space.
LTS~\cite{jing2021locate} innovatively decouples it into a "locate-then-segment" scheme. By explicitly modeling the location's prior knowledge, LTS gets better segmentation performance than the previous best results. CGFormer~\cite{cgformer} proposed a new mask classification framework based on Transformer, which obtains object-level information through mark-based query and grouping strategy and realizes cross-modal and cross-hierarchy reasoning of object perception.

\subsection{Multi-task Visual Grounding (MTVG)}
It is easy to find that, there is a certain correlation between REC and RES, they both need to identify the object in the image with the referring expression. 
Therefore, in recent years, some researchers~\cite{su2023vglaw,luo2020multi,zhu2022seqtr,liu2023polyformer} consider using the same Backbone and cross-modal interaction modules for visual and linguistic feature extraction and multi-modal fusion between them, and then using different heads for detection and segmentation.

MCN~\cite{luo2020multi} proposes a new multi-task collaborative network, which for the first time attempts the joint learning of REC and RES. MCN maximizes the collaborative learning benefits of both REC and RES tasks by leveraging the characteristics of both. 
SeqTr~\cite{zhu2022seqtr} presents a novel approach for MTVG, defining it as a point prediction problem and proposing an innovative, general-purpose network. 
Which unifies disparate visual base tasks under a unified point prediction paradigm, demonstrating the potential for generalization across tasks without modification.
Similar to SeqTr, PolyFormer~\cite{liu2023polyformer} proposes a sequence-to-sequence framework that naturally merges multi-modal features as input sequences and multitask predictions as output sequences. 
In addition, PVD~\cite{cheng2023pvd} has developed a Parallel Vertex Diffusion solution based on the parallelizability of diffusion models. This innovative approach allows for the accurate and efficient generation of vertices in a parallel and scalable manner, offering a competitive advantage in terms of both precision and efficiency.

\section{Method}
\label{method}

In this section, we will provide a detailed introduction to the proposed Progressive Language-guided Visual Learning (PLVL) framework for Multi-Task Visual Grounding (MTVG). Specifically, we first present an overview of the entire framework. Next, we delve into two core designs: the progressive language-guided visual learning backbone and the collaborative multi-task head for the joint Referring Expression Comprehension (REC) and Referring Expression Segmentation (RES). Finally, we outline the training objectives.

\subsection{Overview}

For MTVG, the two sub-tasks REC and RES share the same input, which consists of an image and a referring expression, but produce different outputs, \emph{i.e.}, a bounding box and a segmentation mask, respectively. 
Fig.~\ref{fig:plvl} illustrates the proposed framework, which consists of three parts, \emph{i.e.}, linguistic backbone, language-guided visual backbone, and a collaborative multi-task Head. Given a referring expression and an input image, we first use the linguistic backbone to obtain the language tokens of the 
expression cue. Then, the language tokens and the input image are fed into the visual backbone to extract the useful visual tokens for the subsequent predictions (\emph{i.e.}, bounding box for REC and segmentation mask for RES).

Specifically, to achieve accurate predictions according to the referring expression, the key point is to extract effective visual features which represent to-be-identified objects and match the expression cues. Based on such understanding, we propose a Progressive Language-guided Visual Learning backbone which not only finely mines the inherent feature expression of the visual modality itself but also progressively injects the language expression information to help learn language-related visual features. Besides, we analyze that the predictions of REC and RES share the same center for identifying the object. Unlike conventional approaches which use two independent heads, we propose a collaborative multi-task head that serves as a bridge between REC and RES, enhancing the prediction accuracy of both. Below we will provide the details.

\subsection{Progressive Language-guided Visual Learning}

\begin{figure}[t] 
    \centering  
    \includegraphics[width=1\linewidth]{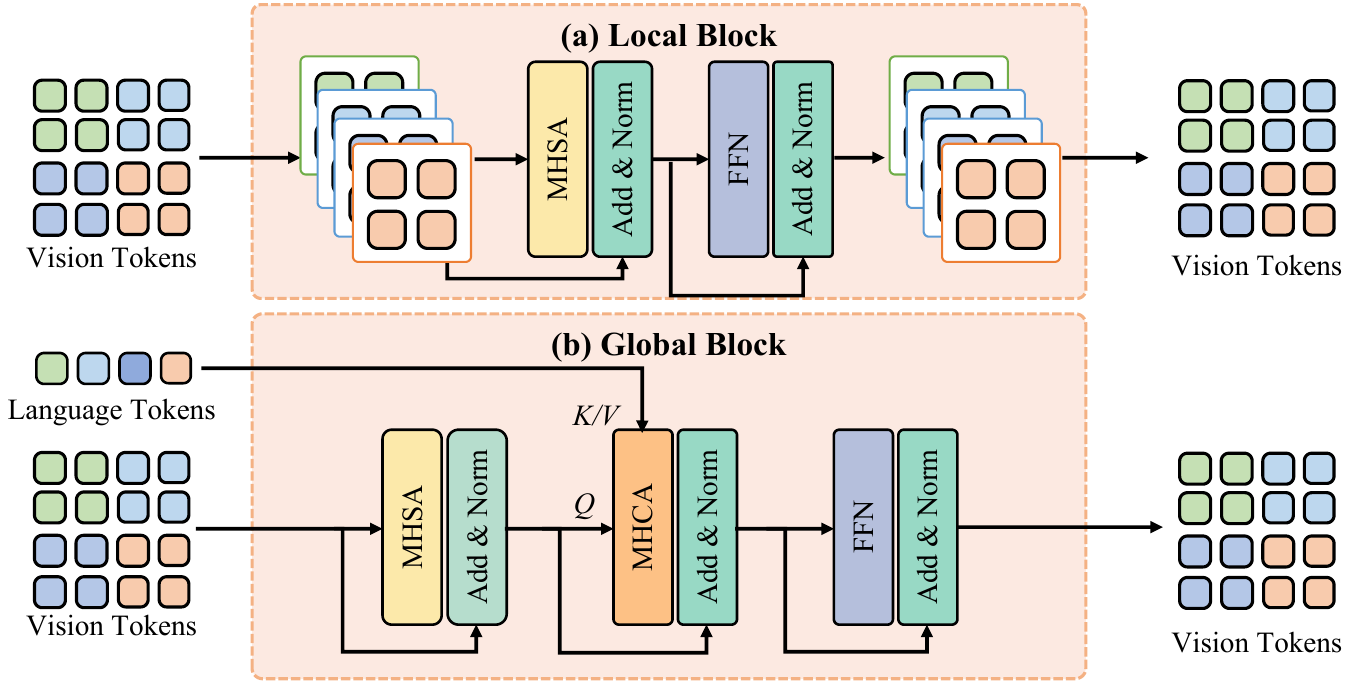}  
    \vspace{-6mm}
    \caption{The structures of local block and global block.}       
    \label{fig:fusion} 
    \vspace{-2mm}
\end{figure}

It is easily understood that for accurate localization and segmentation prediction of to-be-identified objects, the effective feature learning on the image modality itself and the language-relevant feature learning are two key aspects we need to focus on for MTVG. To this end, driven by the excellent capability of ViTDet~\cite{li2022vitdet} in visual feature representation, we propose a local-global-group-based Progressive Language-guided Visual Learning (PLVL) backbones. Please note that the key characteristic of our PLVL lies in that on the basis of the visual backbone of ViTDet, we carefully incorporate the guidance of language information for visual feature learning.

Specifically, as displayed in Fig.~\ref{fig:plvl}, given the referring expression, we follow EEVG~\cite{chen2024efficient} and adopt the 12-layer Bert-Base as the linguistic backbone to extract language tokens $T_{l} \in \mathbb{R}^{N_l \times D}$. 
For the input image $I \in \mathbb{R}^{H \times W \times 3}$, we use the patch Embedding layer to tokenize it as ${T_{v}} \in \mathbb{R}^{H/P \times W/P \times D}$, where $P$ is the Patch size. 
Then, we feed the visual token $T_{v}$ and language token $T_{l}$ into the visual backbone, which sequentially consists of four local-global groups and three global blocks. 
For every local-global group, it is composed of two local blocks and one global block. 
Through the local block, we employ the self-attention mechanism to mine the relationship among the visual modality itself for useful feature representation. 
Through the global block, we progressively inject the guidance of language tokens into the visual backbone via the cross-attention-based interaction mechanism to help learn the effective visual features which match the referring expression. 
In such a progressive learning manner, it is expected that our framework can effectively learn useful visual features for boosting accurate predictions.

Specifically, for the local block, as shown in Fig.~\ref{fig:fusion}(a), in order to reduce the computational complexity, we first split the vision tokens ${T_{v}} \in \mathbb{R}^{H/P \times W/P \times D}$ into four parts as ${T^{i}_{v}} \in \mathbb{R}^{\times H/{2P} \times W/{2P} \times D}$, $i=1,2,3,4$, and feed each part into the same computation procedure consisting of multi-head self-attention (MHSA) and FFN. Then by merging the four computed results along the spatial dimension, we can get the visual features output by the local block. Correspondingly, the entire computation process is formulated as:
\begin{equation}\label{eq2}
\begin{split}
    &[{T^{1}_{v}},{T^{2}_{v}},{T^{3}_{v}},{T^{4}_{v}}]=SS({T_{v}}),\\
    &{{\bar{T}^{i}_{v}}}={{T^{i}_{v}}}+\mathrm{MHSA}({{T^{i}_{v}}}), i=1,2,3,4, \\
    &{{\tilde{T}^{i}_{v}}}={{\bar{T}^{i}_{v}}}+\mathrm{FFN}({{\bar{T}^{i}_{v}}}), \\
    &{{\hat{T}_{v}}}=SC([\tilde{T}^1_{v},\tilde{T}^2_{v},\tilde{T}^3_{v},\tilde{T}^4_{v}]),
\end{split}
\end{equation}
where $SS(\cdot)$ and $SC(\cdot)$ represent the split and concatenation at the spatial dimension, respectively. The local self-attention for every head in $\mathrm{MHSA(\cdot)}$ is defined as:
\begin{equation}\label{sim}
\begin{split}
    {Y} & =\mathrm{Softmax}(\frac{QK^\mathsf{T}}{\sqrt{D}})V,\\
    {Q} & = \phi_Q({{T^{i}_{v}}}), 
    {K} = \phi_K({{T^{i}_{v}}}), 
    {V} = \phi_V({{T^{i}_{v}}}), 
\end{split}
\end{equation}
where $\phi_Q(\cdot)$, $\phi_K(\cdot)$, and $\phi_V(\cdot)$ are linear projection operations for obtaining the query, key, and value, respectively. 

For global block, as illustrated in Fig.~\ref{fig:fusion}(b), unlike the local block, we do not divide the visual tokens spatially. The reason is that during the visual feature learning, it is very important to model the relationship among different division parts. Hence, we directly feed the entire feature $\hat{T}_{v}$ in Eq.~\eqref{eq2} output by local block into a multi-head self-attention (MHSA) procedure. Then, we introduce a multi-head cross-attention (MHCA) module to inject the linguistic information into the backbone for fully extracting the language-related visual information. The complete computation process is:
\begin{equation}\label{eq3}
\begin{split}
    &{{\tilde{T}_{v}}}={{\hat{T}_{v}}}+\mathrm{MHSA}({{\hat{T}_{v}}}), \\
    &{{\bar{T}_{v}}}={{\tilde{T}_{v}}}+\mathrm{MHCA}({{\tilde{T}_{v}}},{T_{l}}), \\
    &{{\bar{\bar{T}}_{v}}}={{\bar{T}_{v}}}+\mathrm{FFN}({{\bar{T}_{v}}}), 
\end{split}
\end{equation}
where the output ${{\bar{\bar{T}}_{v}}}$ is used as the input for the next local block and is fed into the computation procedure in Eq.\eqref{eq2}. The attention operation for every head in $\mathrm{MHCA(\cdot)}$ is:
\begin{equation}\label{sim}
\begin{split}
    {Y} & =\mathrm{Softmax}(\frac{QK^\mathsf{T}}{\sqrt{D}})V,\\
    {Q} & = \phi_Q({{T_{v}}}), 
    {K} = \phi_K({T_{l}}), 
    {V} = \phi_V({T_{l}}).
\end{split}
\end{equation}

As clarified before, the proposed progressive language-guided visual learning backbone is indeed built based on ViTDet~\cite{li2022vitdet}. However, for the multi-task visual grounding task, we have specific and key contributions: 1) Through the progressive integration of the cross-attention mechanism into different global blocks in ViTDet, we carefully inject the linguistic information to help guide the visual feature learning. Such a simple adjustment on the original structure of ViTDet brings obvious performance gains for MTVG, as validated in Table~\ref{tab:backbone} below; 2) We specifically investigate the relationship among REC and RES and construct a collaborative multi-task head, which will be described below.

\subsection{Collaborative Multi-task Head}
\begin{figure} [t] 
    \centering  
    \includegraphics[width=1\linewidth]{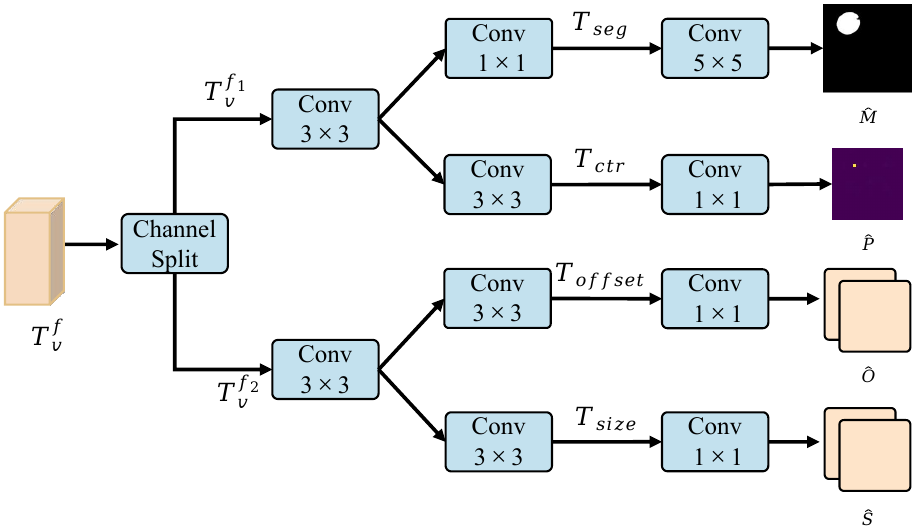}  
    \vspace{-7mm}
    \caption{The structure of collaborative multi-task Head.}       
    \label{fig:head}  
    \vspace{-2mm}
\end{figure}

For the multi-task visual grounding, the existing methods~\cite{chen2024efficient} generally adopt the multi-layer perception (MLP) to construct two independent task-specific heads for separately predicting the bounding box for the REC task and the segmentation mask for the RES task. Although there are some attempts to connect these two prediction procedures via different strategies, such as vertex generation~\cite{zhu2022seqtr,cheng2023pvd}, there is still room for further performance improvement. In this section, we explore the relationship between REC and RES, and construct a novel convolution-based collaborative multi-task head.

Specifically, we rethink the joint prediction of REC and RES, and analyze that the predictions for these two sub-tasks have the similar central position of to-be-identified objects. Since RES is a pixel-wise classification task, if we can build the bridge between the localization center of the REC task and the prediction mask of the RES task, it is expected to better boost the 
mutual predictions for these two sub-tasks. 
Besides, as we know, convolution-based network structures inherently introduce the bias prior, which is beneficial for capturing the locally consistent information of images and then helps the regression prediction. Motivated by these understandings, we build a collaborative multi-task head on the basis of convolution layers, as illustrated in Fig.~\ref{fig:head}. For the REC task, three outputs are predicted, including the coarse-grained prediction of the to-be-identified object's center position, the offsets of the object's real center position from the center position; and the object's size (width and height). For the RES task, we need to classify each vision token and then expand each token into an image patch to accomplish the final mask prediction.

Specifically, given the final visual tokens ${T^{f}_{v}} \in \mathbb{R}^{H/P \times W/P \times D}$ output by the visual backbone, we divide it into two elements ${T^{f_{1}}_{v}} \in \mathbb{R}^{H/P \times W/P \times D/2}$ and ${T^{f_{2}}_{v}} \in \mathbb{R}^{H/P \times W/P \times D/2}$ along the channel dimension. By separately feeding them into a convolutional layer and then adopting different $3\times 3$ convolutional layers for further information propagation, we can obtain the outputs as $T_{seg}$, $T_{ctr}$, $T_{offset}$, and $T_{size}$, respectively.
For the RES task, we input $T_{seg}$ to a convolution layer with a convolution kernel of $5\times 5$ and then obtain the segmentation mask $\hat{M} \in \mathbb{R}^{H \times W}$. Similarly, for the REC task, through the convolutional operations as shown in Fig.~\ref{fig:head}, we can correspondingly get the probability map $\hat{P} \in \mathbb{R}^{H/P \times W/P}$, offset $\hat{O} \in \mathbb{R}^{H/P \times W/P \times 2}$, and the size \textbf{$\hat{S} \in \mathbb{R}^{H/P \times W/P \times 2}$}.
Finally, the bounding box $\hat{B}=(\hat{x},\hat{y},\hat{w},\hat{h})$ is predicted as:
\begin{equation}\label{sim}
\begin{split}
   & \hat{x} = x_d+\hat{O}[x_d,y_d,1],~~~ \hat{y} = y_d+\hat{O}[x_d,y_d,2],\\
    &\hat{w} = \hat{S}[x_d,y_d,1], ~~~\hat{h} = \hat{S}[x_d,y_d,2],
\end{split}
\end{equation}
where $(x_d,y_d)=\underset{(x,y)}{argmax}\hat{P}[x,y]$, $x=1,\ldots, H/P, y=1,\ldots,W/P$. $[x,y]$ represents the element index.

\subsection{Training Objectives}
During experiments, our proposed PLVL framework is directly trained in an end-to-end manner. For REC and RES, the corresponding training objectives $\mathcal{L}_{det}$ and $\mathcal{L}_{seg}$ are:

\begin{equation}
\begin{split}
   & \mathcal{L}_{det} = \mathcal{L}_{focal}(\hat{P}, P) + \mathcal{L}_{1}(\hat{B}, B) + \mathcal{L}_{giou}(\hat{B}, B)), \\
    &    \mathcal{L}_{seg} = \mathcal{L}_{focal}(\hat{M}, M) + \mathcal{L}_{dice}(\hat{M}, M),
    \end{split}
\end{equation}
where $\mathcal{L}_{focal}$ represents focal loss~\cite{lin2017focal}, $\mathcal{L}_{1}$ represents L1 Loss, $\mathcal{L}_{giou}$ represents GIoU loss~\cite{rezatofighi2019generalized}, and $\mathcal{L}_{dice}$ represents dice loss~\cite{milletari2016v}. $B$ and $M$ are the ground-truth bounding box and segmentation mask, respectively.
The label $P$ is generated based on ground-truths.

Finally, the total training loss for MTVG is adopted as:
\begin{equation}
    \mathcal{L} = \lambda_{det}\mathcal{L}_{det} + \lambda_{seg}\mathcal{L}_{seg},
\end{equation}
where $\lambda_{det}$ and $\lambda_{seg}$ are weights used to balance the two losses. Following EEVG~\cite{chen2024efficient}, in this paper, we set $\lambda_{det}=0.1$ and $\lambda_{seg}=1$.

\section{Experiments}
In this section, we first provide detailed experimental settings and then conduct comprehensive comparisons and a series of ablation studies to evaluate the effectiveness of our proposed PLVL.

    \begin{table*}[!t]
\centering

\caption{Comparison with state-of-the-art methods on RefCOCO~\cite{refcoco_and_plus}, RefCOCO+~\cite{refcoco_and_plus}, and RefCOCOg~\cite{nagaraja2016modeling} for \textbf{REC} task. We highlight the best results of traditional setting and pre-trained setting in \textcolor{blue}{blue} and \textcolor{red}{red}. 
}\vspace{-3mm}
\begin{tabular}{c|c|c|c|ccc|ccc|cc}
\hline
\multicolumn{1}{c|}{\multirow{2}{*}{Method}} &\multicolumn{1}{c|}{\multirow{2}{*}{Venue}} & \multicolumn{1}{c|}{\multirow{2}{*}{Backbone}} & \multirow{2}{*}{Multi-task} &\multicolumn{3}{c|}{RefCOCO} & \multicolumn{3}{c|}{RefCOCO+} & \multicolumn{2}{c}{RefCOCOg} \\
& & & &val & test A & test B & val & test A & test B & val(U) & test(U) \\
\hline
\multicolumn{11}{l}{\textbf{Traditional setting}} \\
MCN~\cite{luo2020multi} & CVPR2020 & DarkNet53&\Checkmark & 80.08 & 82.29 & 74.98 & 67.16 & 72.86 & 57.31 & 66.46 & 66.01\\
LBYL~\cite{huang2021look} & CVPR2021 & DarkNet53&\XSolidBrush & 79.67 & 82.91 & 74.15 & 68.64 & 73.38 & 59.49 & - & - \\
TransVG~\cite{deng2021transvg} & ICCV2021 & ResNet101&\XSolidBrush & 81.02 & 82.72 & 78.35 & 64.82 & 70.70 & 56.94 & 68.67 & 67.73 \\
 TRAR~\cite{zhou2021trar} & ICCV2021 & DarkNet53&\XSolidBrush & - & 81.40 & 78.60 & - & 69.10 & 56.10 & 68.90 & 68.30 \\
SeqTR~\cite{zhu2022seqtr} & CVPR2022 & DarkNet53&\Checkmark & 81.23 & 85.00 & 76.08 & 68.82 & 75.37 & 58.78 & {71.35} & {71.58} \\
CLIP-VG~\cite{clipvg} & TMM2023 & ViT-B &\XSolidBrush & 84.29 & 87.76 & 78.43 & 69.55 & 77.33 & 57.62 & 73.18 & 72.54 \\
VG-LAW~\cite{su2023vglaw} & CVPR2023 & ViT-B&\Checkmark & 86.62 & 89.32 & 83.16 & 76.37 & 81.04 & 67.50 & 76.90 & 76.96 \\
ScanFormer~\cite{scanformer} & CVPR2024 & ViLT &\XSolidBrush & 83.40 & 85.86 & 79.81 & 72.96 & 77.57 & 62.50 & 74.10 & 74.14 \\
PVD~\cite{cheng2023pvd} & AAAI2024 & Swin-B &\Checkmark & 84.52 & 87.64 & 79.63 & 73.89 & 78.41 & 64.25 & 73.81 & 74.13 \\
SegVG~\cite{segvg} & ECCV2024 & DETR-Res101 &\XSolidBrush & 86.84 & 89.46 & 83.07 & 77.18 & 82.63 & 67.59 & 78.35 & 77.42 \\
HiVG~\cite{hivg} & ACMMM 2024 & ViT-B &\XSolidBrush & 87.32 & 89.86 & 83.27 & 78.06 & 84.81 & 68.11 & 78.29 & 78.79 \\
EEVG~\cite{chen2024efficient} & ECCV2024 & ViT-B&\Checkmark & 88.08 & \textcolor{blue}{\textbf{90.33}} & 85.50 & 77.97 & 82.44 & 69.15 & 79.60 & 80.24 \\
\textbf{PLVL} & \textbf{Ours} & \textbf{ViT-B} &\Checkmark & \textcolor{blue}{\textbf{89.02}} & {{90.21}} & \textcolor{blue}{\textbf{86.72}} & \textcolor{blue}{\textbf{79.19}} & \textcolor{blue}{\textbf{84.97}} & \textcolor{blue}{\textbf{71.55}} & \textcolor{blue}{\textbf{81.58}} & \textcolor{blue}{\textbf{81.16}} \\
\hline
\multicolumn{11}{l}{\textbf{Pre-trained setting}} \\
RefTr~\cite{li2021referring} & NeurIPS2021 & ResNet101 & \Checkmark & 85.65 & 88.73 & 81.16 & 77.55 & 82.26 & 68.99 & 79.25 & 80.01 \\
SeqTR~\cite{zhu2022seqtr} & CVPR2022 & DarkNet53 & \Checkmark & 87.00& 90.15 & 83.59 & 78.69 & 84.51 & 71.87 & 82.69 & 83.37 \\
PolyFormer~\cite{liu2023polyformer} & CVPR2023 & Swin-B & \Checkmark & 89.73 & 91.73 & 86.03 & 83.73 & 88.60 & 76.38 & 84.46 & 84.96 \\
EEVG~\cite{chen2024efficient} & ECCV2024 & ViT-B & \Checkmark & 90.47 & 92.73 & 87.72 & 81.79 & 87.80 & 74.94 & 85.19 & 84.72 \\
HiVG~\cite{hivg} & ACMMM 2024 & CLIP-B & \XSolidBrush & 90.56 & 92.55 & 87.23 & 83.08 & 89.21 & 76.68 & 86.52 & 56.62 \\
OneRef~\cite{oneref} & NeurIPS2024 & BEiT3-B & \Checkmark & 91.89 & 94.31 & 88.58 & 86.38 & 90.38 & 79.47 & 86.82 & 87.32 \\
\textbf{PLVL} & \textbf{Ours} & \textbf{ViT-B} & \Checkmark & \textcolor{red}{\textbf{92.65}} & \textcolor{red}{\textbf{94.72}} & \textcolor{red}{\textbf{89.95}} & \textcolor{red}{\textbf{87.36}} & \textcolor{red}{\textbf{91.04}} & \textcolor{red}{\textbf{81.60}} & \textcolor{red}{\textbf{89.20}} & \textcolor{red}{\textbf{89.80}} \\

\hline

\end{tabular}

\label{tab:main_rec_det}
\end{table*}

    \begin{table*}[t]
\centering


\caption{Comparison with state-of-the-art methods on RefCOCO~\cite{refcoco_and_plus}, RefCOCO+~\cite{refcoco_and_plus}, and RefCOCOg~\cite{nagaraja2016modeling} for \textbf{RES} task. We highlight the best results of traditional setting and pre-trained setting in \textcolor{blue}{blue} and \textcolor{red}{red}.
}\vspace{-3mm}

\begin{tabular}{c|c|c|c|ccc|ccc|cc}
\hline
\multicolumn{1}{c|}{\multirow{2}{*}{Method}} & \multicolumn{1}{c|}{\multirow{2}{*}{Venue}} & \multicolumn{1}{c|}{\multirow{2}{*}{Backbone}} & \multirow{2}{*}{Multi-task} & \multicolumn{3}{c|}{RefCOCO} & \multicolumn{3}{c|}{RefCOCO+} & \multicolumn{2}{c}{RefCOCOg} \\

 & & & & val & test A & test B & val & test A & test B & val(U) & test(U) \\

\hline
\multicolumn{11}{l}{\textbf{Traditional setting}} \\
MCN~\cite{luo2020multi} & CVPR2020  & DarkNet53&\Checkmark & 62.44 & 64.20 & 59.71 & 50.62 & 54.99 & 44.69 & 49.22 & 49.40  \\
CRIS~\cite{wang2022cris} &CVPR2022 & CLIP-ResNet50&\XSolidBrush & 69.52 & 72.72 & 64.70 & 61.39 & 67.10 & 52.48 & 59.87 & 60.36 \\ 
SeqTR~\cite{zhu2022seqtr} & CVPR2022 & DarkNet53&\Checkmark & 67.26 & 69.79 & 64.12 & 54.14 & 58.93 & 48.19 & 55.67 & 55.64 \\
LAVT~\cite{yang2022lavt} & CVPR2022 & Swin-B &\XSolidBrush & 74.46 & 76.89 & 70.94 & 65.81 & 70.97 & 59.23 & 63.34 & 63.62  \\
VG-LAW~\cite{su2023vglaw} & CVPR2023 & ViT-B&\Checkmark & 75.62 & 77.51 & 72.89 & 66.63 & 70.38 & 59.89 & 65.53 & 66.08 \\
PVD~\cite{cheng2023pvd} & AAAI2024 & Swin-B&\Checkmark & 74.82 & 77.11 &  69.52 & 63.38 & 68.60 & 56.92 & 63.13 & 63.62  \\
EEVG~\cite{chen2024efficient}  & ECCV2024 & ViT-B &\Checkmark  & 78.23 & 79.27 & 76.58 & 69.04 & 72.65 & 62.33 & 69.15 & 70.01 \\
\textbf{PLVL} & \textbf{Ours} & \textbf{ViT-B} &\Checkmark & \textcolor{blue}{\textbf{78.91}} & \textcolor{blue}{\textbf{79.79}} & \textcolor{blue}{\textbf{77.61}} & \textcolor{blue}{\textbf{69.79}} & \textcolor{blue}{\textbf{74.02}} & \textcolor{blue}{\textbf{63.03}} & \textcolor{blue}{\textbf{70.21}} & \textcolor{blue}{\textbf{70.26}} \\

\hline

\multicolumn{11}{l}{\textbf{Pre-trained setting}} \\
RefTr~\cite{li2021referring} & NeurIPS2021 & ResNet101 & \Checkmark & 74.34 & 76.77 & 70.87 & 66.75 & 70.58 & 59.40 & 66.63 & 67.39 \\
SeqTR~\cite{zhu2022seqtr} & CVPR2022 & DarkNet53 & \Checkmark & 71.70 & 73.31 & 69.82 & 63.04 & 66.73 & 58.97 & 64.69 & 65.74 \\
PolyFormer~\cite{liu2023polyformer} & CVPR2023 &Swin-B & \Checkmark & 75.96 & 77.09 & 73.22 & 70.65 & 74.51 & 64.64 & 69.36 & 69.88 \\
EEVG~\cite{chen2024efficient}  & ECCV2024 & ViT-B & \Checkmark & 79.49 & 80.87 & 77.39 & 71.86 & 76.67 & 66.31 & 73.56 & 73.47 \\
OneRef~\cite{oneref} & NeurIPS2024 & BEiT3-B & \Checkmark & {79.83}  & {81.86}  & {76.99}  & {74.68}  & {77.90}  & {69.58}  & {74.06}  & {74.92} \\
\textbf{PLVL} & \textbf{Ours} & \textbf{ViT-B} & \Checkmark & \textcolor{red}{\textbf{81.89}} & \textcolor{red}{\textbf{82.95}} & \textcolor{red}{\textbf{80.04}} & \textcolor{red}{\textbf{76.68}} & \textcolor{red}{\textbf{79.67}} & \textcolor{red}{\textbf{72.65}} & \textcolor{red}{\textbf{77.25}} & \textcolor{red}{\textbf{77.67}} \\
\hline
\end{tabular}
\label{tab:main_res_seg}
\end{table*}

    
\subsection{Experimental Settings}

\noindent\textbf{Datasets.}
We conduct experiments on three mainstream benchmark datasets, including RefCOCO, RefCOCO+~\cite{refcoco_and_plus}, and RefCOCOg~\cite{mao2016generation}. They are all collected from MS-COCO~\cite{lin2014microsoft}.
RefCOCO contains 19,994 images and 142,210 reference expressions to represent 50,000 objects, which is divided into a validation set, a test set A, and a test set B.
RefCOCO+ contains 19,992 images and 141,564 reference expressions to represent 49,856 objects, divided into a validation set, a test set A, and a test set B.
RefCOCOg contains 25,799 images and 95,010 reference expressions to represent 49,822 objects, divided into a validation set, and a test set. 

\vspace{1mm}
\noindent\textbf{Evaluation Metrics.} Following most existing work, for REC, we adopt the classic 
intersection over union (IoU) as the evaluation metric with the threshold as 0.5. For RES, we use the mean intersection over union (mIoU) between the predicted result and the ground-truth mask as an evaluation metric. 

\vspace{1mm}
\noindent\textbf{Implementation Details.}
Our experiments are implemented based on PyTorch by using 4 NVIDIA A100 GPUs. 
The model is end-to-end optimized by AdamW~\cite{adamw} and the weight decay is $1\times10^{-4}$. 
The resolution of the input image is resized to 448 × 448 and the referring expressions are padded or truncated to 40 tokens. 
The initial learning rate of the backbone encoder is $5\times10^{-6}$, the initial learning rate of the self-attention and FFN in the first 12 layers of the decoder is $1\times10^{-5}$, and the initial learning rate of the rest of the model is $2.5\times10^{-5}$.
We use the pre-trained BERT-base to initialize the linguistic backbone, the pre-trained ViTDet~\cite{li2022vitdet} to initialize the self-attention and FFN in the first 12 layers of the visual backbone, and use a uniform distribution to initialize the parameters of the rest of the model.
We use distributed training, with 20 pairs of samples on each GPU, a total batch size of 80, and a total epochs of 150.

\begin{figure*}[t] 
    \centering  
    \includegraphics[width=1\linewidth]{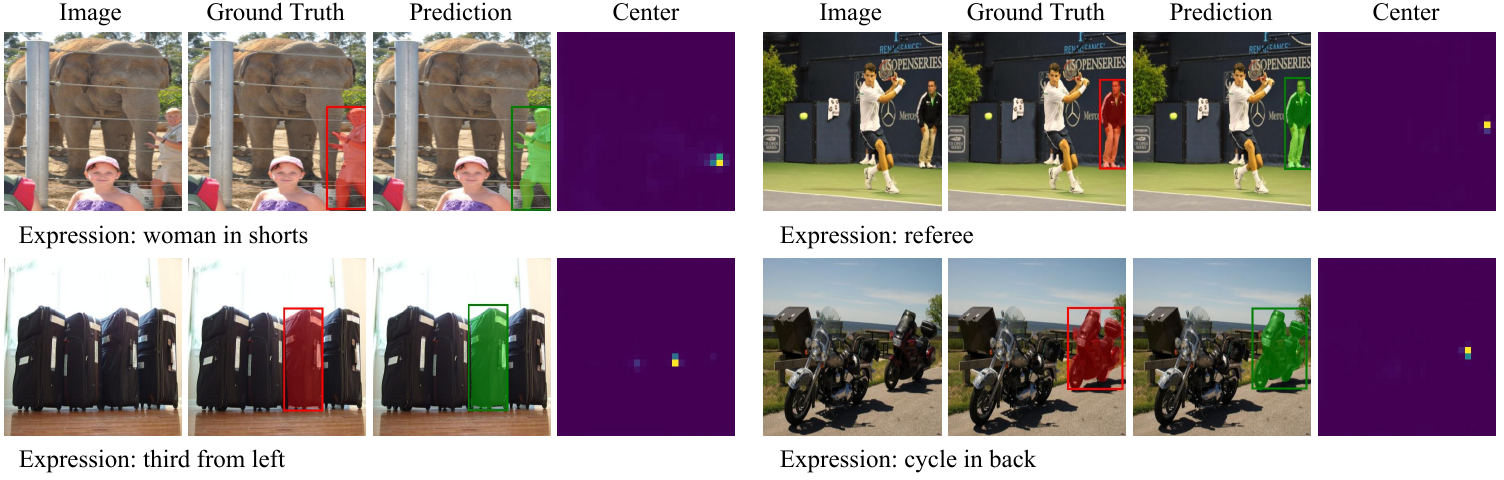}  
    \vspace{-6mm}
    \caption{Qualitative results on the RefCOCO. From left to right: the input image, the ground truth of REC and RES, the predicted results of PLVL, the score map of REC sub-task.}       
    \label{fig:vis_refcoco} 
\end{figure*}

\begin{figure*}[t] 
    \centering  
    \includegraphics[width=1\linewidth]{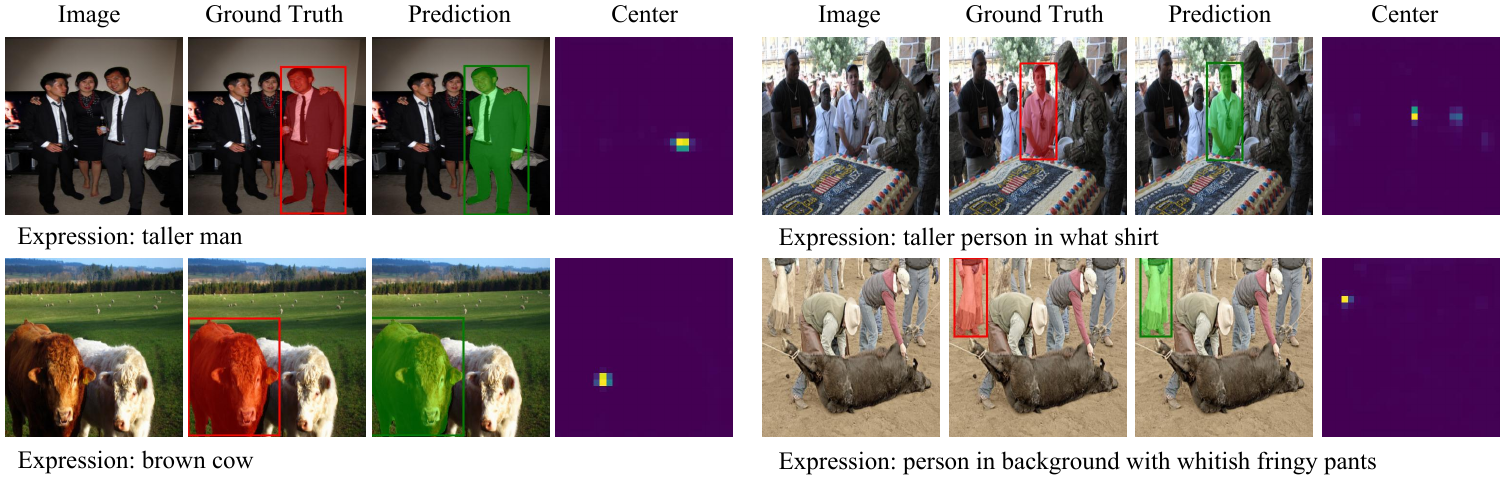}  
      \vspace{-6mm}
    \caption{Qualitative results on the RefCOCO+. From left to right: the input image, the ground truth of REC and RES, the predicted results of PLVL, the score map of REC sub-task.}       
    \label{fig:vis_refcoco+} 
\end{figure*}

\subsection{Experimental Comparisons}
Here we conduct comprehensive comparisons based on traditional setting and pre-trained setting where different methods are first trained on a large corpus of visual grounding data which consists of RefCOCO, RefCOCO+, RefCOCOg, Visual Genome~\cite{krishna2017visual} and Flickr30k~\cite{plummer2015flickr30k} and then fine-tuned on the benchmark dataset.

\vspace{1mm}
\noindent\textbf{Evaluation under Traditional setting.} For REC and RES tasks, the quantitative results under the conventional training setting are reported in the upper part of Table ~\ref{tab:main_rec_det} and the upper part of Table ~\ref{tab:main_res_seg}, respectively. As seen, for these two different tasks, our proposed PLVL consistently achieves superior performance on these three benchmark datasets. These results finely substantiate the rationality of our proposed language-guided progressive visual learning mechanism and the multi-task collaborative prediction strategy.

\begin{figure*}[] 
    \centering  
    \includegraphics[width=1\linewidth]{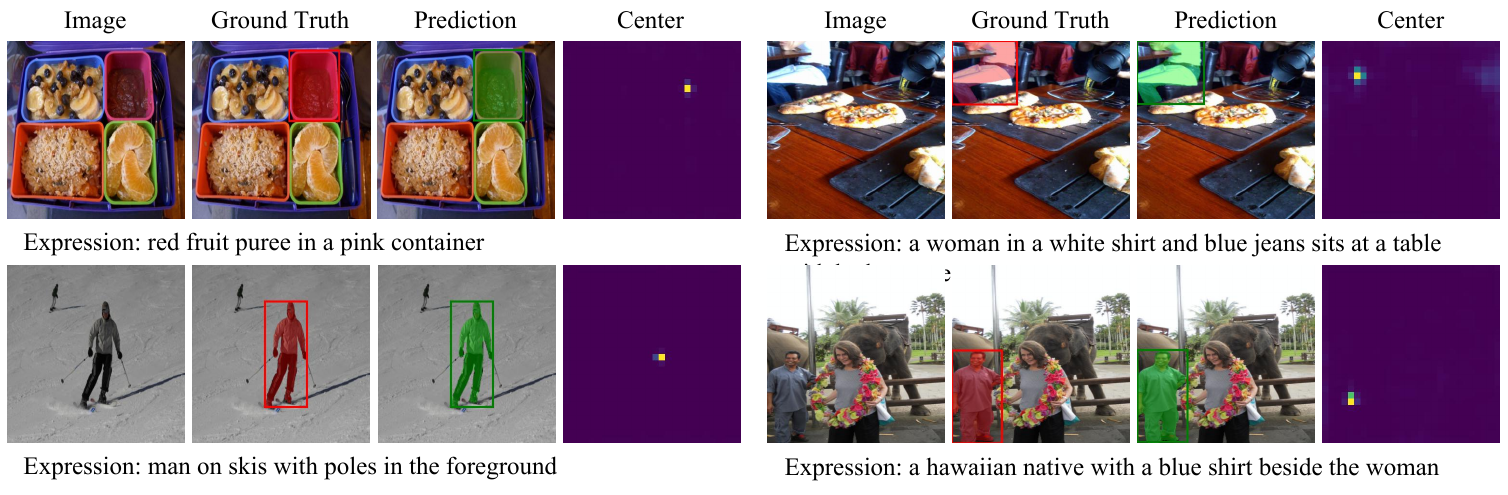}
      \vspace{-6mm}
    \caption{Qualitative results on the RefCOCOg. From left to right: the input image, the ground truth of REC and RES, the predicted results of PLVL, the score map of REC sub-task.}       
    \label{fig:vis_refcocog} 
\end{figure*}

 Fig.~\ref{fig:vis_refcoco}, Fig.~\ref{fig:vis_refcoco+}, and Fig.~\ref{fig:vis_refcocog} present the visual results on different samples which are randomly selected from these three benchmark datasets. As observed, our proposed method can detect and segment the referred objects accurately. 
 Besides, we can see that the maximum response location in the score map output by the center branch finely aligns with the center of the mask, which complies with the design motivation of the multi-task head.


\vspace{1mm}
\noindent\textbf{Evaluation under Pre-trained setting.} For comprehensive comparison, we provide the results of different methods on the REC and the RES tasks under the pre-trained setting, which are listed in the lower part of Table~\ref{tab:main_rec_det} and the lower part of
Table~\ref{tab:main_res_seg}, respectively.
We can see that the pre-trained procedure assists different approaches in achieving better performance, and our PLVL consistently outperforms other baselines on different datasets.


\vspace{1mm}
\noindent\textbf{Comparisons on Inference Time.}
Table~\ref{tab:speed_sota} reports the comparison with existing state-of-the-art (SOTA) open-sourced methods in terms of inference time and computational cost. Especially, compared to the latest SOTA EEVG, we reduce the time overhead by 8\% per image. As seen,
our PLVL has much lower computational cost and fewer network parameters.



\begin{table}[t]
\centering
\caption{Computational cost, inference time, and number of parameters of different state-of-the-art methods. ``$\downarrow$'' means lower is better. The batch size is 1 and all experiments are conducted in an A100 GPU. }\vspace{-3mm}
\begin{tabular}{c|ccc}
\noalign{\hrule height 1.5pt} 
Method & PolyFormer~\cite{liu2023polyformer} & EEVG~\cite{chen2024efficient} & PLVL \\
\hline
GPLOPs $\downarrow$       & -     & 75.22G & 75.04G \\
\hline
Runtime (ms) $\downarrow$ & 82.72 & 50.04  & 45.96 \\
\hline
\#Param & 309.73M & 218.07M & 218.02M \\
\noalign{\hrule height 1.5pt}
\end{tabular}
\label{tab:speed_sota}
\vspace{-3mm}
\end{table}


\subsection{Ablation Studies}
Based on RefCOCOg, we conduct a series of ablation studies to validate the effectiveness of our proposed method, including the progressive injection of linguistic information,  the cooperative potential of REC and RES, and the collaborative multi-head paradigm.

\vspace{1mm}
\noindent\textbf{Effectiveness of Language-guided Visual Learning.}
From Fig.~\ref{fig:plvl}, the visual backbone of our PLVL consists of 8 local blocks and 7 global blocks with progressively injecting the language guidance information through the cross-attention module.
To verify the effectiveness of the proposed progressive visual learning, Table~\ref{tab:backbone} reports the performance with adopting different numbers of global blocks. We can find that 
a decrease in cross-attention modules before 12 layers leads to a notable drop in performance for REC and RES. This finding supports the notion that the full use of language information can enhance the extraction of visual features.
\begin{table}[t]
\centering
\caption{Performance on RefCOCOg validation set with adopting different numbers of global blocks. $N$ is the number of local blocks, $M$ is the number of global blocks with language guidance, and Indexes represent the indexes where we insert the language information into the global blocks.}
\vspace{-3mm}
\begin{tabular}{ccc|cc}
\noalign{\hrule height 1.5pt}
{$N$} & {$M$} & {Indexes} & {REC} & {RES} \\
\noalign{\hrule height 1.5pt} 
12 & 3 & 13,14,15                & 80.15       & 69.51      \\
11 & 4 & 3,13,14,15              & 80.02       & 69.81      \\
10 & 5 & 3,6,13,14,15            & 80.78       & 69.24      \\
9  & 6 & 3,6,9,13,14,15          & 81.50       & 70.16      \\
8  & 7 & 3,6,9,12,13,14,15       & 81.58       & 70.21      \\
\noalign{\hrule height 1.5pt}

\end{tabular}
\label{tab:backbone}
\end{table}

\vspace{1mm}
\noindent\textbf{Effectiveness of Multi-Task Strategy.}
We compare the performance of the single-task networks with our proposed multitask strategy on the RefCOCOg validation set. Table~\ref{tab:multi_task} shows that our multitask strategy can bring significant performance gains of 1.43\% on REC and 1.13\% on RES.
These results demonstrate that a well-designed multi-task approach can leverage the task relationship to further improve the performance of individual tasks.

\begin{table}[t]
\centering
\caption{Performance comparison between multi-task and single-task for training on RefCOCOg validation set.}
\vspace{-3mm}
\begin{tabular}{cc|cc}
\noalign{\hrule height 1.5pt}
{$\mathcal{L}_{det}$} & {$\mathcal{L}_{seg}$} & {REC} & {RES} \\
\noalign{\hrule height 1.5pt}
\Checkmark   & \Checkmark   & 81.58  & 70.21  \\
\Checkmark   & \XSolidBrush & 80.15  & -      \\
\XSolidBrush & \Checkmark   & -      & 69.08  \\
\noalign{\hrule height 1.5pt}

\end{tabular}
\vspace{-2mm}
\label{tab:multi_task}
\end{table}


\vspace{1mm}
\noindent\textbf{Effectiveness of Collaborative Task Grouping.}
We compare different collaborative task groupings for RES and REC. Table~\ref{tab:head} indicates that grouping $\{mask, center\}$ in one group and $\{offset, size\}$ in another leads to a 0.21\% performance gain for the REC task and a 0.37\% performance gain for the RES task. The experimental results also demonstrate that the design of multi-task grouping can further enhance the model's performance.

\begin{table}[t]
\centering
\caption{Performance comparison between different head types on RefCOCOg validation set.}
\vspace{-3mm}
\begin{tabular}{c|cc}
\noalign{\hrule height 1.5pt}
Head type & {REC} & {RES} \\
\noalign{\hrule height 1.5pt} 
\{mask\}\{center,offset,size\} & 81.37  & 69.84 \\
\{mask,center\}\{offset,size\} & 81.58  & 70.21 \\
\noalign{\hrule height 1.5pt}
\end{tabular}
\label{tab:head}
\end{table}

\noindent\textbf{Flexibility of our proposed PLVL and CMTH.}
In order to further verify the advantages of our proposed core contribution points, we incorporate our proposed Progressive Language-guided Visual Learning (PLVL) and Collaborative Multi-task Head (CMTH) into EEVG. 
As shown in Table~\ref{tab:eevg}, our proposed PLVL and CMTH can both bring significant performance improvements on EEVG.
These experimental results show that the two core contributions we propose have the potential to be applied to other architectures.

\begin{table}[t]
\centering
\caption{Performance comparison between EEVG w. or w/o PLVL and CMTH.}
\vspace{-3mm}
\begin{tabular}{c|cc}
\noalign{\hrule height 1.5pt}
Head type & {REC} & {RES} \\
\noalign{\hrule height 1.5pt} 
EEVG  & 79.27  & 68.16 \\
EEVG w. CMTH & 79.98  & 68.66 \\
EEVG w. PLVL & 80.35  & 69.48 \\
EEVG w. PLVL \& CMTH & 81.41  & 69.78 \\
\noalign{\hrule height 1.5pt}

\end{tabular}
\vspace{-3mm}
\label{tab:eevg}
\end{table}

\section{Conclusion}
In this paper, for multi-task visual grounding, we have proposed a novel Progressive Language-guided Visual Learning (PLVL) framework.
Not only does it mine the inherent feature representations of the visual modality itself, but it also progressively integrates linguistic information to enhance the learning of language-related visual features.
Furthermore, we have carefully investigated the relationship between the two tasks of REC and RES and proposed a collaborative multi-task head to enhance the performance of both tasks. Through extensive experiments on several benchmark datasets, the effectiveness of our proposed method has been validated, demonstrating its superiority over existing techniques in terms of both REC and RES. 



\bibliographystyle{ACM-Reference-Format}
\bibliography{authordraft}










\end{document}